\documentclass{article}

\usepackage{arxiv}

\usepackage{makecell}
\usepackage[utf8]{inputenc} 
\usepackage[T1]{fontenc}    
\usepackage{hyperref}       
\usepackage{url}            
\usepackage{booktabs}       
\usepackage{amsfonts}       
\usepackage{nicefrac}       
\usepackage{microtype}      
\usepackage{graphicx}
\usepackage{doi}
\usepackage[numbers]{natbib}
\usepackage{setspace}
\usepackage{amsmath}
\usepackage{amssymb}
\usepackage{eucal}
\usepackage{tcolorbox}
\tcbset{
  myquestionbox/.style={
    colback=gray!5,
    colframe=black,
    boxrule=0.5pt,
    arc=2pt,
    left=4pt,
    right=4pt,
    top=4pt,
    bottom=4pt,
    fonttitle=\bfseries,
    before skip=8pt,
    after skip=8pt
  }
}
\usepackage[export]{adjustbox}
\usepackage{multirow}
\usepackage{booktabs}  
\usepackage{rotating}
\usepackage{soul,xcolor}

\usepackage{algorithm}
\usepackage{algpseudocode}
\usepackage{listings}
\usepackage{xspace}

\def\tsc#1{\csdef{#1}{\textsc{\lowercase{#1}}\xspace}}
\tsc{WGM}
\tsc{QE}
\tsc{EP}
\tsc{PMS}
\tsc{BEC}
\tsc{DE}

\title{Hyperdimensional Computing for Sustainable Manufacturing: An Initial Assessment}

\date{} 					

\author{Danny Hoang\\
	School of Mechanical, Aerospace, and Manufacturing Engineering\\
	University of Connecticut\\
	Storrs, Connecticut, USA\\
	\And
Anandkumar Patel \\
	Department of Mechanical and Aerospace Engineering\\
    Rutgers University\\
    New Brunswick, NJ, USA
	\AND
Ruimen Chen\\
	School of Mechanical, Aerospace, and Manufacturing Engineering\\
	University of Connecticut\\
	Storrs, Connecticut, USA\\
    \AND
Rajiv Malhotra\\
    Department of Mechanical and Aerospace Engineering\\
    Rutgers University\\
    New Brunswick, NJ, USA\\
	\And
	Farhad Imani \thanks{Corresponding author: farhad.imani@uconn.edu}\\ School of Mechanical, Aerospace, and Manufacturing Engineering\\
	University of Connecticut\\
	Storrs, Connecticut, USA\\
}

\renewcommand{\headeright}{}
\renewcommand{\undertitle}{}
\renewcommand{\shorttitle}{Hyperdimensional Computing for Sustainable Manufacturing}

\hypersetup{
pdftitle={Hyperdimensional Computing for Sustainable Manufacturing: An Initial Assessment},
pdfauthor={Danny Hoang, Anandkumar Patel, Rajiv Malhotra, Farhad Imani},
pdfkeywords={Smart Manufacturing, Artificial Intelligence, Knowledge Hyperdimensional Computing
}
}

\begin{document}
\maketitle
\begin{abstract}
Smart manufacturing can significantly improve efficiency and reduce energy consumption, yet the energy demands of AI models may offset these gains. This study utilizes in-situ sensing-based prediction of geometric quality in smart machining to compare the energy consumption, accuracy, and speed of common AI models. HyperDimensional Computing (HDC) is introduced as an alternative, achieving accuracy comparable to conventional models while drastically reducing energy consumption, 200$\times$ for training and 175 to 1000$\times$ for inference. Furthermore, HDC reduces training times by 200$\times$ and inference times by 300 to 600$\times$, showcasing its potential for energy-efficient smart manufacturing.

\end{abstract}

\section{Introduction}
Smart manufacturing has shown the potential to reduce energy consumption and enhance sustainability by increasing material and operational efficiency~\cite{matsunaga2022optimization,sahoo2022smart}. A critical enabler of these benefits is the integration of in-situ sensing and analysis, which provides real-time insights into manufacturing processes to allow data-driven adjustments. This capability complements three key aspects of smart manufacturing where Artificial Intelligence (AI) plays a key role, i.e., modeling and optimization under limited mechanistic knowledge~\cite{li2023deep,madni2019leveraging}, process and equipment analysis and monitoring~\cite{gao2022guest}, and real-time process and system control~\cite{chen2022transfer,fernandes2022machine,razvi2019review,wang2020machine,tercan2022machine}. The as-yet limited adoption of smart manufacturing by small and medium enterprises, which comprise $\approx$ 75\% of US manufacturing, indicates tremendous growth potential.

Research in computer science and sustainability indicates that training and inference of AI models needs significant energy ~\cite{nishant2020artificial,chen2023survey}. Thus, the sustainability advantage of smart manufacturing could be negated by energy-hungry training and inference of the requisite AI models. Optimization of computational workloads only reduces training costs by 10s of percentage points, often at the cost of reduced training/inference speed or prediction accuracy~\cite{yang2023chasing,you2023zeus}. This is problematic for smart manufacturing, e.g., rapid and repeated inference is critical for real-time control and offline optimization. The cumulative energy usage of repeated inference can overwhelm that of training, which is significantly less frequent~\cite{desislavov2023trends}. At the same time, speedy training is critical when adapting pre-trained AI models to unseen scenarios (e.g., new materials or part geometries). 

In this context, in-situ sensing presents an opportunity in smart manufacturing to reduce computational energy demands by providing real-time data that enables more efficient AI pipelines. The integration of sensing technologies provides the ability to mitigate energy usage without compromising prediction accuracy or training and inference speeds. Therefore, there is a need to quantify the energy usage of AI models in smart manufacturing, more specifically in scenarios that emphasize in situ sensing and analysis, to realize sustainable and efficient manufacturing operations. 

This paper presents our early, but promising, work towards addressing this need. In-situ monitoring of geometric deviation in 5-axis Computer Numerical Control (CNC) machining is used as an example testbed. Multiple real-time acquired sensor signals from the CNC are fed to an AI model that classifies geometric deviation of relevant features as low, average or high. Several popular AI models and the rarely explored HyperDimensional Computing (HDC) approach are examined. In addition to comparing accuracy and training/inference speed a key novelty lies in comparison of the energy needed for training and inference. The future impact on balancing pervasiveness of AI-driven smart manufacturing with sustainability is discussed.

\begin{figure*}
    \centering
    \includegraphics[width=\textwidth]{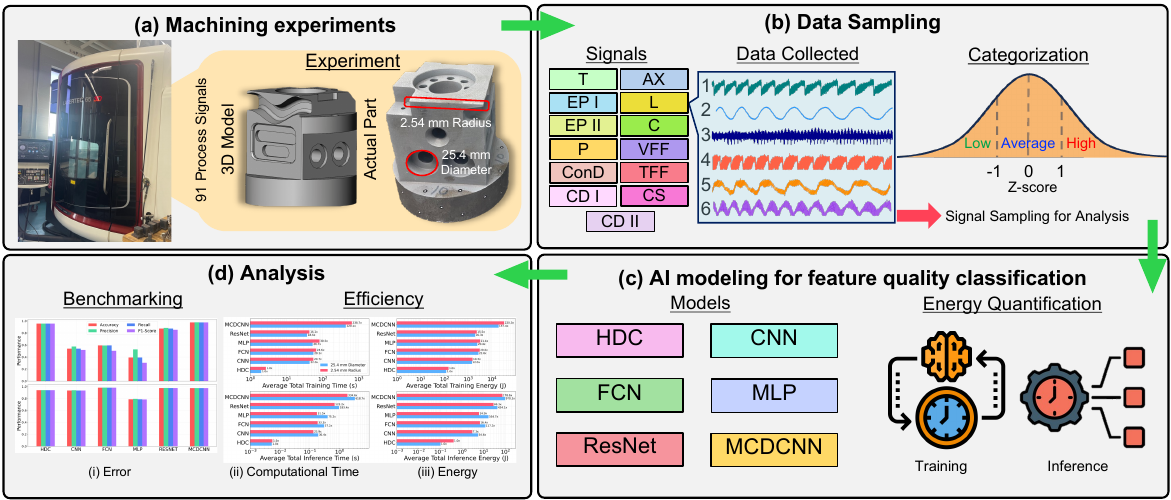}
    \vspace{-5mm}
    \caption{Flowchart of the overall approach adopted in this paper.}
    \label{flowchart}
\end{figure*}

\section{Methodology}
Figure \ref{flowchart} shows a schematic of the methodology. It consists of three elements, i.e., experiments and data generation, AI modeling and comparison of different leaning methods.

\subsection{Experiments}\label{experiments_methodology}
Training and testing data were experimentally generated using a LASERTEC 65 DED hybrid CNC DMG machine to fabricate the complex part depicted in Fig. \ref{flowchart}a. A total of 18 such parts were machined using 1040 Steel feedstock and same machine settings. Worn bits were replaced based on expert experience. During fabrication, signals from the CNC were recorded at a frequency of 500 Hz using a Siemens Simatic IPC227E device. The toolpaths were created using Siemens NX. Two features were selected for analysis, i.e., a 25.4 mm left counterbore diameter and a 2.54 mm radius, as highlighted in red in Fig. \ref{flowchart}a. The counterbore was created using an end mill bit, and the radius was generated using a corner radius tool.

The real-time measured signals consisted of load (L), current (C), torque (T), command speed (CS), control differential I (CD I), control differential II (CD II), power (P), contour deviation (ConD), axis position (AX), encoder position I (EP I), encoder position II (EP II), velocity feed forward (VFF), and torque feed forward (TFF). The signals CD I, CD II, EP I, and EP II are proprietary to the machine. Each signal was recorded on either five channels (axes only) or six channels (axes plus spindle), across the five axes and the spindle, resulting in a total of 64 channels. Sampling was performed specifically on the recorded channels to ensure sufficient data for training purposes. The sampling involves non-overlapping n-gram windows that capture specific time intervals in the time series data. These windows are concatenated to create a single feature vector sample, increasing the number of data samples while preserving the temporal relationship. Each class was balanced. A GOM ATOS ScanBox was used to measure the geometry of each machined feature and compared it to the ideal geometry. Z-scores were computed for each of these geometric deviations and grouped into three categories: scores below -1 indicating low, scores between -1 and 1 indicating average, and scores above 1 indicating high levels of deviation (Fig. \ref{flowchart}b).

\subsection{AI modeling methods}
The AI models used temporal signals for each feature as inputs and output the corresponding category of geometric deviation. Several AI models that have been widely used in the manufacturing literature were trained (Fig. \ref{flowchart}c). These included Convolutional Neural Network (CNN)~\cite{mangrolia2024continuing,chen2023transfer}, Fully Convolutional Neural Networks (FCN), Multilayer Perceptron (MLP), Residual Network (RESNET), and multi-channel deep convolutional neural network (MCDCNN). With an 80-20 test-train split, each model was run 50 times using an NVIDIA GeForce RTX 4090 GPU.

In addition, HDC was also used as an AI model. Since HDC is not as common in the manufacturing field as the above AI models, we describe it briefly while also referring the reader to the literature for further details~\cite{kleyko2022survey,hoang2024hierarchical, hoang2025hyperdimensional, chen2025federated}.

Hyperdimensional computing is based on using high dimensional vectors to represent data through a mapping function and subsequently manipulating them for learning. The generated vectors distribute the original information over all the dimensions of the hypervectors such that each dimension is equally important. This holographic reduced representation mimics how the human brain operates, wherein information is not stored in one place but spread across multiple locations. Representing each piece of information as hypervectors allows the use of well-defined operations such as addition, multiplication, and permutation, to augment the data stored within and to learn from the information. Further, the implementation of these operations is highly parallelizable, resulting in faster training and inference compared to traditional gradient-based learning methods.

HDC consists of encoding and learning stages. Encoding creates a hyperdimensional space using orthogonal basis vectors $\mathbf{B} = \{\mathbf{B}_{1},\mathbf{B}_{2},...,\mathbf{B}_{m}\}$, where $\mathbf{B}_{i} \, \forall i \in 1,...,m$ has a dimensionality of $D$. Each element of the basis vector is randomly sampled from a standard normal distribution. The resulting vector is usable for data encoding in high dimensions. Due to the large dimension of the vector and the random sampling from a standard normal distribution two basis vectors exhibit nearly orthogonal behavior. Given a set of input signal vectors denoted as $\mathcal{\mathbf{X}} = \{\mathbf{x}_1, \mathbf{x}_2, ..., \mathbf{x}_N\}$ and their output labels $\mathcal{\mathbf{Y}} = \{y_1, y_2, ..., y_N\}$, each input vector in $\mathcal{\mathbf{X}}$ is mapped to a high-dimensional space, i.e., set of hypervectors denoted as $\mathbf{F} = \{\mathbf{F}_{1}, \mathbf{F}_{2}, ..., \mathbf{F}_{N}\}$, each with a dimension of $D$. \textcolor{black}{In this experiment each $\mathbf{x}_i$ is the concatenated n-gram feature vector over all 64 channels as described in Section \ref{experiments_methodology}, so the HDC encoder maps all signals in the n-gram jointly into a single hypervector, combining information across signals in one representation.}

The information within each hypervector is condensed into distinct class hypervectors based on their labels by bundling (i.e., element-wise addition) the hypervectors in each class. As a result, a collection of class hypervectors is generated, effectively capturing the characteristics associated with each associated input vector and allowing classification of new feature vectors. The encoding efficiently encodes high-dimensional data and extracts key information for classification in the hyperdimensional space. Once encoded hypervectors are generated, training hypervectors corresponding to the same label are bundled together to create class hypervectors corresponding to the dimensional accuracy of workpieces. A linear training element using cosine similarity is incorporated considering all label information in $\mathcal{\mathbf{Y}}$. Cosine similarity $\delta$ is defined as
\begin{equation}
        \delta(\mathbf{H}_1, \mathbf{H}_2) = \frac{ \mathbf{H}_1 \cdot \mathbf{H}_2}{\|\mathbf{H}_1\|  \|\mathbf{H}_2\|}
\end{equation}
where $\mathbf{H_1}$ and $\mathbf{H_2}$ are two hypervectors. Hypervectors that are more closely related have $\delta$ closer to 1 and those that are different have $\delta$ closer to 0. HDC employs iterative retraining by updating the class hypervectors when the current model mispredicts incoming training query data $\mathbf{I}$. If the current iteration predicts a query $c$ as $c'$, the retraining process adjusts the class hypervectors $\mathbf{C}_{c}$ and $\mathbf{C}_{c'}$ as:

\begin{equation}
    \begin{array}{l}
        \mathbf{C}_c \leftarrow \mathbf{C}_c + \eta (1-\delta(\mathbf{C}_c, \mathbf{I})) \times \mathbf{I} \\
        \mathbf{C}_{c'} \leftarrow {\mathbf{C}}_{c'} -\eta (1-\delta(\mathbf{C}_{c'}, \mathbf{I})) \times {\mathbf{I}} 
    \end{array} 
    \label{HDCupdate}
\end{equation}

\noindent where $\eta$ represents a chosen learning rate. As a result of this update scheme, the HDC model demonstrates higher initial accuracy and converges with fewer iterations. In the inference stage an input vector $\mathbf{x_q}$ is mapped using the above discussed encoding to the same $D$ dimension to create the query hypervector $\mathbf{Q}$. The corresponding classification prediction $\hat{y}$ is then expressed as
\begin{equation}
    \hat{y} = \arg\max_{c \in \mathcal{C}} \, \delta(\mathbf{Q}, \mathbf{C}_c)
\end{equation}
where the highest similarity between the class hypervectors is selected as the classification result.

\subsection{Analysis}
The classification ability of the conventional AI models and HDC was compared by quantifying accuracy, precision, recall, and F1-Score. The energy usage of the model for training and inference was measured using the Zeus Python library. \textcolor{black}{This library queries the NVML API to measure GPU execution time and power, and thus reports only GPU-board energy, excluding CPU and host device data transfer overheads. As such, the reported gains characterize relative GPU energy efficiency rather than full system-level energy.} The total energy required to run the code for training or inference is computed by integrating the power usage over the time.

\section{Results}
\begin{figure}[ht!]
    \centering
    \includegraphics[width=\textwidth]{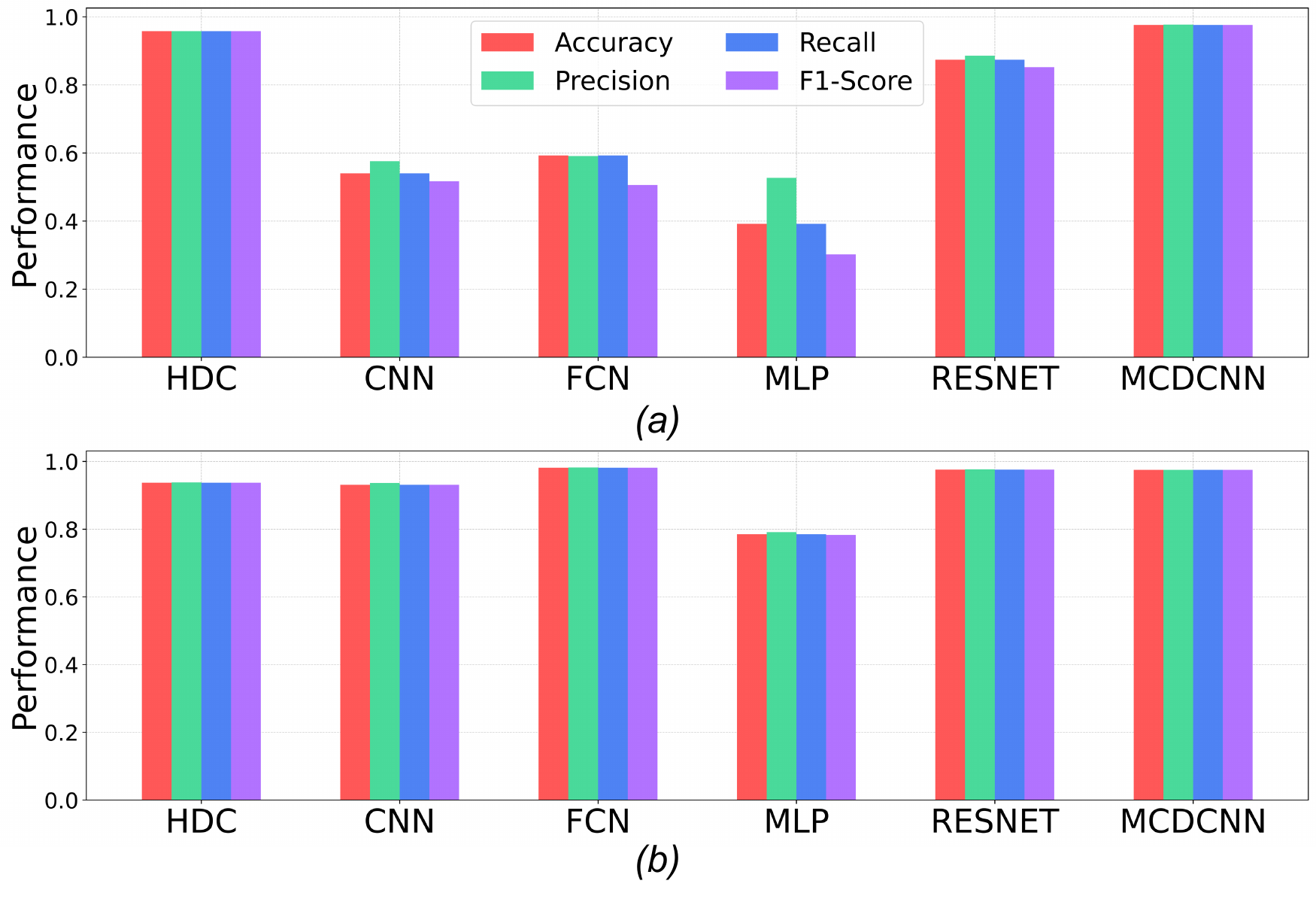}
    \caption{Comparison of HDC to other deep learning models in terms of accuracy, precision, recall, and F1-Score for a) 25.4 mm Diameter feature and b) 2.54 mm radius feature.}
    \label{fig:benchmark}
\end{figure}

\textcolor{black}{Figure \ref{fig:benchmark} compares the performance of HDC, with dimensionality of 10000 and learning rate of 0.05, to the other AI models in terms of accuracy, precision, recall, and F1-Score.} For the 25.4 mm counterbore feature (see Fig. \ref{fig:benchmark}a) HDC has similar performance to MCDCNN, whose performance is in turn higher than that of RESNET, CNN, FCN, and MLP. For the 2.54 mm radius feature (Fig. \ref{fig:benchmark}b) HDC achieves similar performance as other methods with the exception of MLP. Overall, this indicates that HDC and MCDCNN have similar performance across diverse features.

\begin{figure*}
    \centering
    \includegraphics[width=\textwidth]{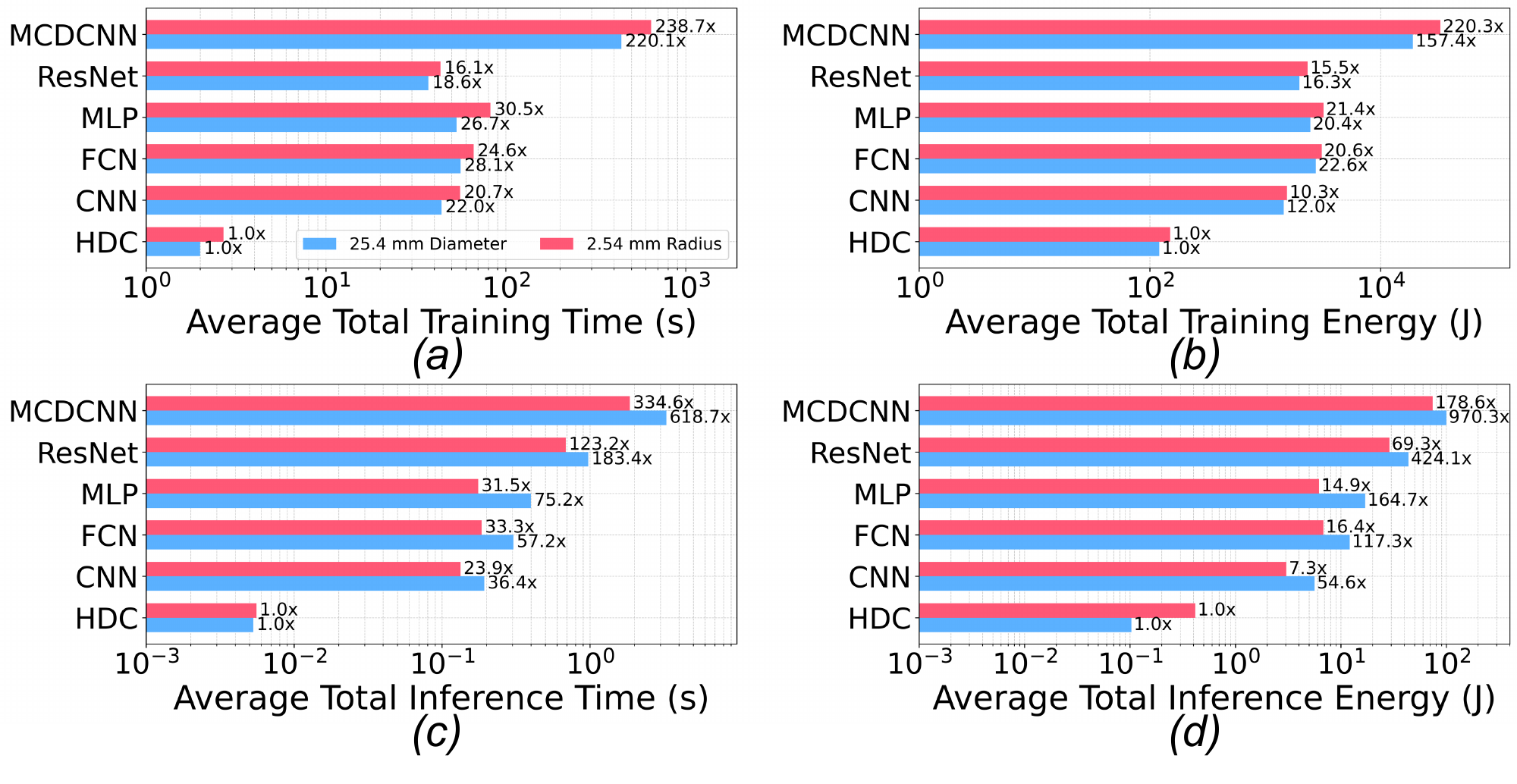}
    \caption{Comparison of HDC to other models. Increase in time and energy efficiency relative to HDC is shown next to each bar. (a) Average total training time and speedup (b) Average total training energy usage (c) Average total inference time and (d) Average total inference energy usage.}
    \label{fig:efficiency}
\end{figure*}

\textcolor{black}{Figure \ref{fig:efficiency} compares the time required and the energy necessary for training and inference. For much larger models such as MCDCNN and RESNET, HDC achieves a maximum decrease in training time by $\approx$ 200$\times$ (Fig. \ref{fig:efficiency}a) and a reduction in inference time by 300-600$\times$ (Fig \ref{fig:efficiency}b). HDC also reduces the energy consumption for training by $\approx$ 200$\times$ and the energy usage for inference by 175 to 1000$\times$. Compared to smaller models such as MLP, FCN, and CNN, performance is still greatly improved with HDC achieving uplifts of at least 20$\times$, 10$\times$, 23$\times$, and 7.3$\times$ for training time, training energy, inference time, and inference energy, respectively. Taken together, these results demonstrate that HDC consistently outperforms both deeper and lighter-weight models, providing orders of magnitude gains in time and energy efficiency for training and inference.}

We further estimate the energy savings possible with HDC in the context of the anticipated proliferation of smart manufacturing. We focus on real-time process control, i.e., acquisition of signals during fabrication and use of AI models for determining quality during rather than after fabrication. This approach is of increasing interest in the manufacturing community since it ensures part quality and high productivity by allowing fabrication to be interrupted if bad parts are being created~\cite{fernandes2022machine,razvi2019review,wang2020machine,tercan2022machine}. Such in-process quality determination can also be used for real-time control to minimize defects or deviations and thus increase yield ~\cite{mangrolia2024continuing,jin2019intelligent,kershaw2021hybrid}. But such real-time monitoring also requires multiple inferences of the trained AI model during the fabrication process. This makes it a good case study for conservatively examining the impact of HDC on energy reduction in smart manufacturing. The global energy usage per annum $E$ by an AI model trained for the above real-time process control can be estimated as
\begin{equation}
    E = E_I \times \nu \times \tau \times \rho \times N_M \times N_P
    \label{global}
\end{equation}
\noindent $E_I$ is the energy used per inference, $\nu$ is the frequency of inference during fabrication, $\tau$ is the part fabrication time, $\rho$ is the number of parts made per year, $N_M$ is the number of machines worldwide for the process of interest (e.g., forming, machining, joining etc.) and $N_P$ is the number of processes.

The following assumptions are made to quantify $E$. The scope
is limited to CNC, i.e., $N_P=1$. The AI model is used for inferencing at a frequency of $\nu=1$ Hz. This is a conservative estimate since the frequency of features in most signals in manufacturing processes is typically higher. The average time taken to machine a part is assumed to be $\tau = 1$ hour. This is also a conservative estimate since the fabrication time for complex multi-feature components, similar to those shown in Fig. \ref{flowchart}a, is usually higher. The number of such parts made per year is assumed to be $\rho=1000$. This amounts to $\approx 4$ parts per working day, assuming a work day of 8 hours equates to a cautiously low machine utilization of 50\%. It is also assumed that the total number of CNC machining centers used worldwide reaches $N_M= 1$ million by 2030 and that they all use the above real-time quality determination capability. Note that market estimates indicate that this estimation of $N_M$ is possibly on the lower end. The energy per inference is based on the computations for HDC and MCDCNN, as shown in Figure \ref{fig:efficiency}, since these two methods yield similar accuracy. Thus, $E_I$ is estimated as $\approx 10^{-1}$ J for HDC and as $\approx 10^1$ J for MCDCNN.

Based on Equation \ref{global} and the above assumptions, the energy saving realized by HDC is $\approx 3.6 \times10^{13}$ Joules or $\approx 10^7$ kWh. As per the US EPA this is equivalent to 7000 tons of $\text{CO}_2$ equivalent or $\text{CO}_2$ emissions from more than 1500 cars driven for a year. Based on the specific energy of manufacturing processes this energy savings corresponds to machining of $\approx 10^4$ cubic feet of Titanium and $\approx 10^7$  kg of stampings~\cite{balogun2016specific}. Note that this estimate is only for one process type and only for real-time quality estimation. Increasing the number of processes beyond machining, expanding the use of AI for inference towards real-time process and system control and machine maintenance, and retraining of AI models for different manufacturing scenarios, will increase these substantial energy and $\text{CO}_2$ savings even more. 

\section{Conclusions}
This paper presents a comparative analysis of various AI models for smart manufacturing in the context of in-situ monitoring of geometric deviations in 5-axis CNC machining as a testbed. The results show that HyperDimensional Computing (HDC) achieves similar prediction accuracy as traditional AI models, but significantly reduces the energy and time needed for training and inference.
Specifically, HDC reduces training time by 200$\times$ and inference time by 300-600$\times$ as compared to conventional AI models. Moreover, HDC also realizes an energy reduction of 200$\times$ for training and 175 to 1000$\times$ for inference. Conservative assessment of the resulting industry-wide impact in future consisting of pervasive smart manufacturing indicates the potential of HDC to realize significant energy savings and $\text{CO}_2$ reduction. Future work will focus on expanding HDC to other processes and exploring its integration into closed-loop control, part design, and machine tool maintenance.

\section*{Declaration of Competing Interest}
The authors declare that they have no known competing financial interests or personal relationships that could have appeared to influence the work reported in this paper.

\section*{Acknowledgment}
The authors gratefully acknowledge the valuable contributions from the Connecticut Center for Advanced Technology (CCAT) to this research. This work was supported by grants 2434519, 2434385, 2146062, and 2001081 from the U.S. National Science Foundation, and by the U.S. Department of Energy under grant number DE-EE0011029.


\bibliographystyle{model1-num-names}

\bibliography{asmejour-sample}


\end{document}